\documentclass[conference]{IEEEtran}
\IEEEoverridecommandlockouts
\usepackage{cite}
\usepackage{amsmath,amssymb,amsfonts}
\usepackage{graphicx}
\usepackage{textcomp}
\usepackage{xcolor}
\def\BibTeX{{\rm B\kern-.05em{\sc i\kern-.025em b}\kern-.08em
    T\kern-.1667em\lower.7ex\hbox{E}\kern-.125emX}}
    
\usepackage{url}
\usepackage{amsmath}
\usepackage{color}

\usepackage[normalem]{ ulem }
\usepackage{soul}
\usepackage{xcolor}
\usepackage{verbatim}
\usepackage[colorinlistoftodos,prependcaption,textsize=tiny]{todonotes}
\newcommand{\ak}[1]{\textcolor{black}{#1}}

\DeclareMathOperator*{\argmin}{arg\,min}
\usepackage{algorithm}
\usepackage{algpseudocode}
\DeclareMathSymbol{*}{\mathbin}{symbols}{"03}
\renewcommand{\algorithmicrequire}{\textbf{Input:}}
\renewcommand{\algorithmicensure}{\textbf{Output:}}

\usepackage{array,multirow,makecell}
\newcolumntype{R}[1]{>{\raggedleft\arraybackslash }b{#1}}
\newcolumntype{L}[1]{>{\raggedright\arraybackslash }b{#1}}
\newcolumntype{C}[1]{>{\centering\arraybackslash }b{#1}}

\begin{document}

\title{Mapping individual differences in cortical architecture using multi-view representation learning}

\author{\IEEEauthorblockN{Akrem Sellami \textsuperscript{1,2}, François-Xavier Dupé \textsuperscript{1}, Bastien Cagna \textsuperscript{2}, Hachem Kadri \textsuperscript{1}, Stéphane Ayache \textsuperscript{1}, Thierry Artières \textsuperscript{1,3}, \\ and Sylvain Takerkart \textsuperscript{2}}
\IEEEauthorblockA{
\textit{\textsuperscript{1}Aix Marseille Université, Université de Toulon, CNRS, LIS, Marseille, France}\\
\textit{\textsuperscript{2}Institut de Neurosciences de la Timone UMR 7289, Aix-Marseille Université, CNRS, Marseille, France}\\
\textit{\textsuperscript{3}École Centrale de Marseille}\\
$\{$akrem.sellami, francois-xavier.dupe, hachem.kadri, stephane.ayache, thierry.artieres$\}$@lis-lab.fr \\
$\{$bastien.cagna, sylvain.takerkart$\}$@univ-amu.fr }
}

\maketitle

\begin{abstract}
In neuroscience, understanding inter-individual differences has recently emerged as a major challenge, for which functional magnetic resonance imaging (fMRI) has proven invaluable. For this, neuroscientists rely on basic methods such as univariate linear correlations between single brain features and a score that quantifies either the severity of a disease or the subject's performance in a cognitive task. However, to this date, task-fMRI and resting-state fMRI have been exploited separately for this question, because of the lack of methods to effectively combine them. In this paper, we introduce a novel machine learning method which allows combining the activation- and connectivity-based information respectively measured through these two fMRI protocols to identify markers of
individual differences in the functional organization of the brain. It combines a  multi-view deep autoencoder which is designed to fuse the two fMRI modalities into a joint representation space within which
a predictive model is trained to guess a scalar score that characterizes the patient.
Our experimental results demonstrate the ability of the proposed method to outperform competitive approaches and to produce interpretable and biologically plausible results.
\end{abstract}


\section{Introduction}
Since its discovery in the 1990s, functional magnetic resonance imaging (fMRI) has played a critical role in developing our understanding of the brain and its dysfunctions.
The research based on fMRI has primarily focused on identifying population-wise principles that emerge from averaging observations across a group of subjects.
In cognitive neuroscience, fMRI has \ak{greatly furthered} our knowledge of the organization of the brain of healthy individuals by mapping \textit{brain activity} induced by a set of tasks that the participant performs in a controlled manner within the scanner (task-fMRI).
More recently, the mapping of \textit{functional connectivity} from fMRI recordings of participants lying at rest in the scanner (rest-fMRI) has complemented our understanding of the functional architecture of the brain.
In particular, because of its ease of acquisition, rest-fMRI has become a tool of choice in clinical neuroscience to characterize brain dysfunction in groups of patients.

 Usually, the specificities of the fMRI signatures extracted in each subject are simply regarded as noise.
However, it is now clear that these specificities carry information that can be used to characterize individual differences, measured e.g as the subject's performance in a cognitive task or on a clinical scale that quantifies the severity of disease (see \cite{seghier_interpreting_2018}, and also the full issue of NeuroImage dedicated to this question \cite{calhoun2017prediction}).
The standard methods used by neuroscientists to study individual differences are based on univariate correlational analysis between a chosen brain feature and such a \ak{behavioral score.}
While it is now clearly established that the design of predictive models should be favoured in this context~\cite{dubois_building_2016}, and that multi-modal data fusion provides a strong added value~\cite{calhoun_multimodal_2016}, the question of fusing features extracted from task- and rest-fMRI has not been addressed~\cite{calhoun_multimodal_2016} even though their complementarity is clearly established~\cite{cole_intrinsic_2014}.

In this paper, we introduce a novel machine learning method that exactly aims at combining
the activation- and connectivity-based information estimated from different fMRI protocols to build a predictive model of individual differences.

First, the challenge of combining heterogeneous information from task- and rest-fMRI is met for the first time by capitalizing on the recent advances in unsupervised representation learning, and more precisely  \ak{ by developing} a specific multi-view deep autoencoder. The main goal of multi-view, or multi-modal, representation learning is to  \ak{ learn} a latent space which combines the information from different views available on the same data, where such a latent representation space is assumed to be \ak{informative enough}  to reconstruct the corresponding views~\cite{zhao2017multi,li2018survey}. According to the literature, the most common ways of dealing with multiple views consist in either concatenating the view-specific layers (a direct fusion), sharing a hidden layer~\cite{ngiam2011multimodal,feng2014cross}, (\ak{e.g.} using canonical correlation analysis~\cite{wang2015deep,kan2016multi,somandepalli2019multimodal}) or finding a direct projection from one view to another~\cite{hsu2018unsupervised}. In particular, \ak{the concatenation of layers}  can provide efficient representations of multi-modal data when the relationships between modalities are non-linear~\cite{ngiam2011multimodal}, such as it may be expected with task- and rest-fMRI data.

Secondly, a trace regression model is estimated in this representation space to allow predicting individual characteristics of new patients, measured as their behavioral performance or the severity of their pathology.
This also yields a mapping of the brain that makes it possible to identify the nodes of the relevant macroscopic cortical network(s).
The experimental validation is conducted on a recently published public dataset that aims at studying individual differences in voice recognition abilities using multi-modal MRI data~\cite{data}. Our experiments quantitatively demonstrate 
the advantages provided by the combination of task- and rest-fMRI when compared to an equivalent model learnt on single-view representations. They also show that our multi-view representation learning frawework yields interpretable results that are consistent with the most recent neuroscientific literature.

The remainder of the paper is organized as follows. In Section~\ref{method}, we detail the proposed approach, including the multi-view autoencoder that allows fusing task- and rest-fMRI data, and the trace regression predictive model. In Section~\ref{results}, we first describe the dataset itself together with the pre-processing pipeline that yields multi-modal features; then we detail our experimental protocol, a quantitative evaluation that demonstrates the superiority of
the multi-view autoencoder and a neuroscientific interpretation that shows
the applicative potential of our method.
We conclude in Section~\ref{conclusion} by discussing opportunities for future work.

\section{Method}
\label{method}
In this section, we detail the different methodological components of the proposed approach. The goal of this study is to design a model able to predict the behavioral score of a subject (e.g. the performance in a voice recognition task) from multi-view fMRI data of this subject. The main aspects of this setting is the multimodal and high dimensional nature of the raw data and the very small number of subjects (usually tens). 

To tackle such a challenge, we choose to design a multimodal feature extractor that combines task- and rest-fMRI point wise (it is learned on isolated vertices of the cortical surface) in order to get enough training material. The fusion of the complementary activation- and connectivity-based information is performed by a deep multi-view autoencoder operating on input pairs of low dimensional feature vectors representing both fMRI modalities for a given vertex. This model is learned on hundreds of thousands of data points.  

\begin{figure*}[!ht]
    \centering
    \includegraphics[scale=0.4]{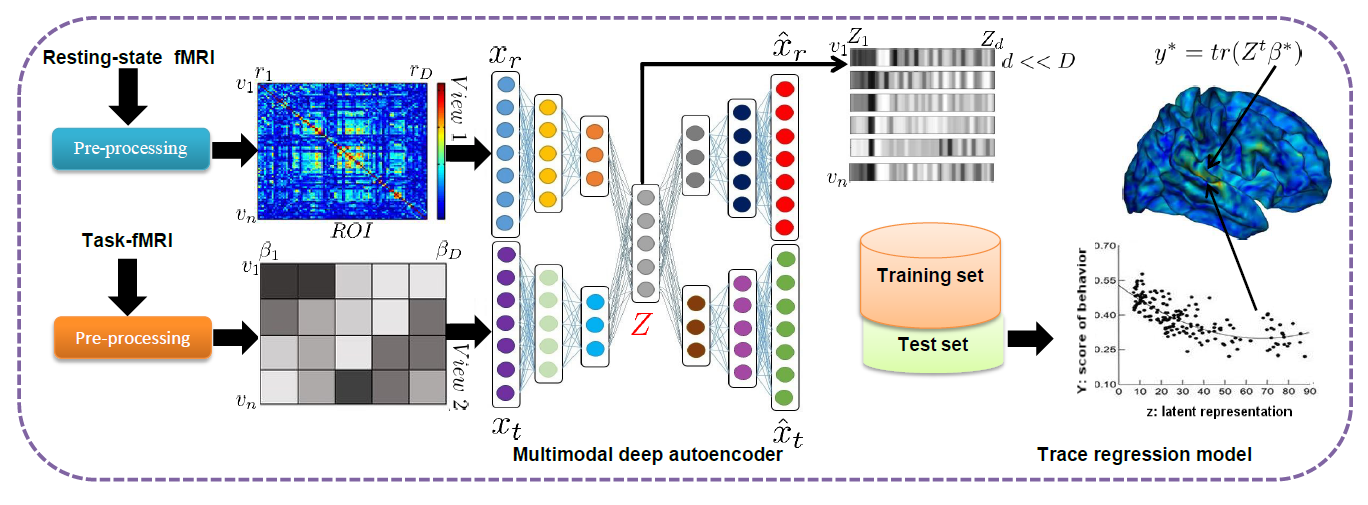}
    \caption{General overview of the proposed machine learning \ak{method} }
    \label{proposed_approach}
\end{figure*}

This feature extractor is used to project the whole brain of each subject into a new representation space within which a predictive model of inter-individual differences (trained to predict a behavorial score) is learned using a trace regression model estimated in the matrix-valued space generated by the bottleneck layer of the autoencoder. This predictive model operates at the subject level and  is then learned on tens of samples only. 
Figure~\ref{proposed_approach} provides a general overview of this approach.
\subsection{Multimodal deep autoencoder for learning brain features representation}
We designed a multimodal autoencoder that aims at extracting \ak{high} level features from the combination of multiple input modalities, from which these multiple input modalities may be reconstructed \cite{ngiam2011multimodal}.

\begin{figure*}[!ht]
    \centering
    \includegraphics[scale=0.18]{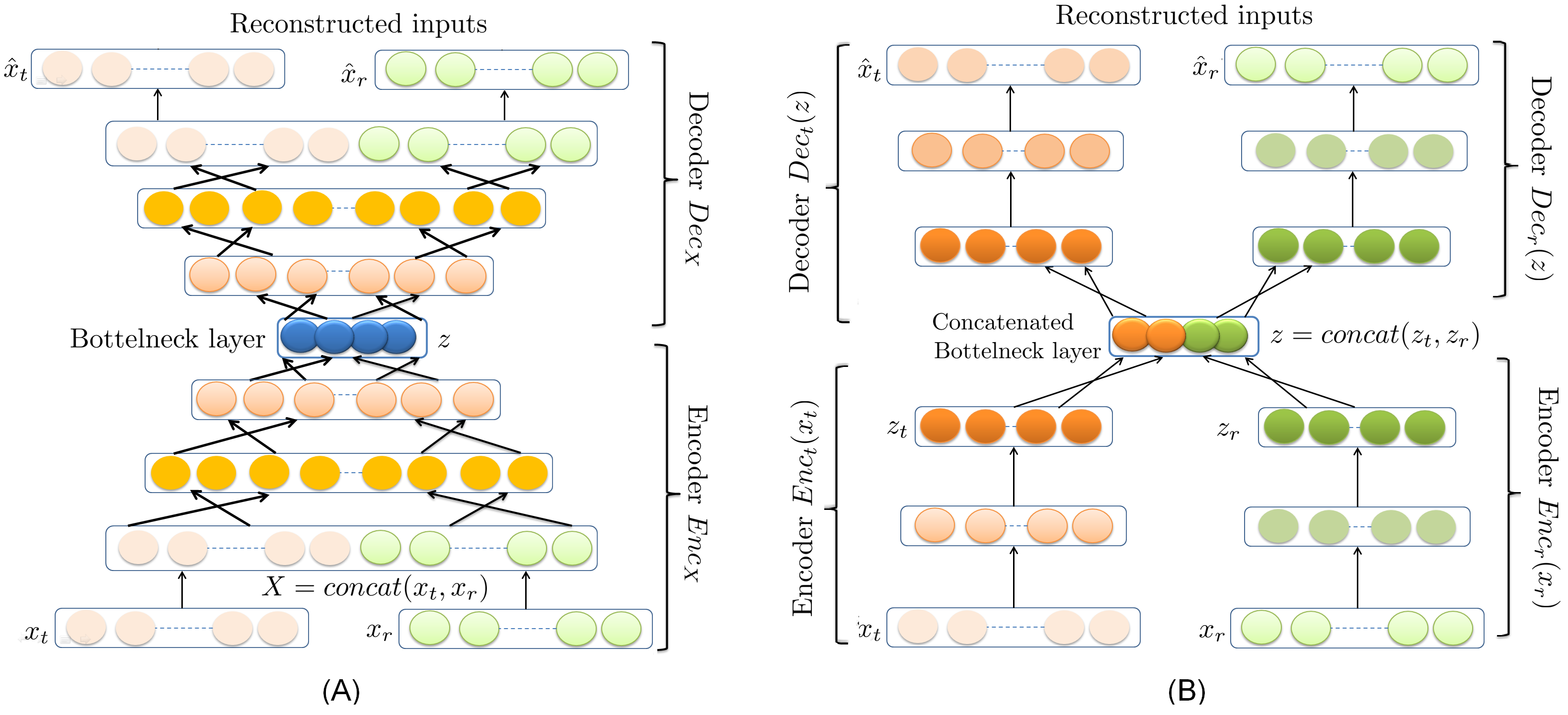}
    \caption{The two architectures considered for the multimodal representation learning scheme. (A) AE, a standard autoencoder trained  on the concatenation of inputs from different modalities. (b) MDAE, a multimodal deep autoencoder where latent representations from the different modalities are concatenated in the bottleneck layer ($z$ is the combination of $z_{r}$ and $z_{t}$). }
    \label{multimodal}
\end{figure*}

We investigated two main autoencoder architectures which are illustrated in Figure \ref{multimodal}. In both cases, the input to the autoencoder is a pair $(x_{r},x_{t})$ of the two MRI modalities (inputs), $x_r$ for rest-fMRI and $x_t$ for task-based fMRI.  

\ak{A first architecture, hereafter simply called autoencoder (and noted AE) consists in a standard autoencoder which takes as input the concatenation of inputs} $x=concat(x_{r},x_{t})$. The AE includes an encoder noted $E_{AE}$ and a decoder which are both implemented as dense multi-layer neural networks (with one to three hidden layers). The encoder non-linearly projects the concatenated inputs into a new representation space $z=E_{AE}(x)$ from which the decoder $D_{AE}$ aims to recover $x$, i.e. $D_{AE}(E_{AE}(x))\approx x$. The autoencoder is classically trained  using a Mean Squared Error (MSE) criterion:

$$\frac{1}{n \times m}\sum_{i=1}^n{\sum_{j=1}^m{\Vert D_{AE}(E_{AE}(x^{i,j})) - x^{i,j}   \Vert^2}}$$ 
where $m$ is the number of vertex of the 3D cortical mesh (see hereafter) and $n$ is the number of subjects, and $x^{i,j}=concat(x^{i,j}_r,x^{i,j}_s)$ is the $({i,j})^{th}$ training sample corresponding to the concatenation of the two fMRI data feature vectors for a given vertex $i$ and for subject $m$, $x_r^{i,j}$ and $x_t^{i,j}$.

Alternatively, we investigated a multi-view deep autoencoder (MDAE) which includes one specific encoder per modality (view), noted $E_r$ and $E_t$ for rest-fMRI and task-based fMRI inputs. The two encoders project each of the modalities into a new representation space $z_r=E_r(x_r)$ and $z_t=E_t(x_t)$ where the representations are concatenated ($z=concat(z_r,z_t)$) and input to two decoders $D_r$ and $D_t$ that aim at recovering the corresponding inputs $x_r$ and $x_t$. Here again all encoders and decoders are implemented as multi-layer neural networks (we used one to three hidden layers) and the training is performed by minimizing a MSE criterion: 
\begin{align*}
    & \frac{1}{n \times m} \sum_{i=1}^n{\sum_{j=1}^m{\Vert D_r(concat(E_r(x_r^{i,j}), E_t(x_t^{i,j}))) - x_r^{i,j}   \Vert^2}} \\
    &+ \frac{1}{n} \sum_{i=1}^n{\sum_{j=1}^m{\Vert D_t(concat(E_r(x_r^{i,j}), E_t(x_t^{i,j}))) - x_t^{i,j}   \Vert^2 }}
\end{align*}

After training, the models estimated using these two architectures may then be used to compute a multi-view encoding of an input pair $(x_r, x_t)$ through $z_{AE} = E_{AE}(concat(x_r,x_t))$ or $z_{MDAE}=concat(E_r(x_r), E_t(x_t))$.

\subsection{Predictive model of individual differences}
 
We now aim at building a model that is able to predict a scalar score associated with each subject, e.g a behavioral score that measures the performance of the subject in a cognitive task, from the multi-view encoding $z$ available at all cortical locations / vertices.
We therefore frame this problem as a regression task. Linear regression has a long history~\cite{friedman2001elements}, from the least-squares algorithm to the Lasso and its extensions. Modern methods combine common regression loss functions with appropriate regularization such as elastic net~\cite{zhang2017discriminative}, (group)-sparsity, manifold~\cite{belkin2006manifold}. Sparsity constraints allow dealing with small sample sizes by seeking for a low dimensional linear subspace. Manifold constraints enforce the solution to belong to a given topology (a mesh for example) encoded by a graph Laplacian. While classical linear regression models seek to predict a scalar response from a vector-valued input, several extensions have been proposed to handle structured inputs such as matrices or tensors~\cite{koltchinskii2011nuclear}. In this case, regularization based on low-rankness or group sparsity are used to capture the structure of the data~\cite{koltchinskii2011nuclear, zhao2017trace}.

In neuroimaging, previous studies (e.g.~\cite{grosenick2013interpretable}) approach the behavioral score prediction problem by learning a standard linear regression model using a training set $\{(z_i,y_i)\}$, $i=1,\ldots,n$, where $z_i \in \mathbb{R}^d $ is a vector of explanatory variables,  $y_i \in \mathbb{R}$ is the output response (i.e., the behavioral score), and $n$ is the number of examples~(subjects).  
In contrast, our approach exploits the specific characteristics of our data. First the multi-view features computed by the deep autoencoder form a matrix of explanatory variables $Z_i \in \mathbb{R}^{m \times d}$, where $m$ is the number of vertices of each subject and $d$ is the number of extracted features (size of the multi-view encoding). In order to handle such matrix-valued inputs, we here consider a {\em trace regression} model. Trace regression is a natural model for matrix-type input data~\cite{koltchinskii2011nuclear, slawski2015regularization, fan2019generalized}. It provides a linear model that maps matrices to real-valued outputs, and is a generalization of the well-studied linear regression model.
The trace regression model corresponds to re-arranging the coefficients in the traditional linear model as a matrix. It is defined as follows
\begin{equation*}
\label{eq:traceRegressionModel}
y_i = \mathrm{tr}(\beta^{\star\top} Z_i) + \epsilon_i, \ i=1,\ldots,n,
\end{equation*}
where $\mathrm{tr}(\cdot)$ denotes the trace, $Z_i\in\mathbb{R}^{m \times d}$ is a matrix-variate input, $\beta^\star\in\mathbb{R}^{m \times d}$ is the matrix of regression coefficients, $y_i$ is the scalar output and $\epsilon_i$ is a random noise.

Moreover, we consider a brain mesh as a graph for designing a specific regularization.  Most of previous studies on trace regression perform estimation of the model parameters via regularized least squares with a matrix norm regularization: 
\begin{equation}
\label{eq:traceRegressionOptim}
\hat{\beta} = \argmin_{\beta\in\mathbb{R}^{m \times d}}\left\{ \sum_{i=1}^n \big(y_i - \mathrm{tr}(\beta^{\top} Z_i)\big)^2 + \Omega(\beta) \right\},
\end{equation}
where $\Omega(\beta)$ is a matrix regularizer which may lead to low-rank or/and structured sparse solutions~\cite{koltchinskii2011nuclear, zhao2017trace, fan2019generalized}. In our case, we want to exploit the knowledge that,  $\forall i, {Z_i}$ are related to vector on a mesh (the brain surface), and it should give some benefit to link $\hat{\beta}$ with the underlying manifold. Following~\cite{grosenick2013interpretable}, we use a manifold regularization which forces each vertex to have feature values close to each other. Formally, given $L$ the Laplacian of the graph of the underlying manifold, we define the regularization term as: $\Omega_1(\beta) = \eta \mathrm{tr}(\beta^\top L \beta)$.
Furthermore, since we want to obtain interpretable weights $\beta$, we add a sparse prior on the vertices such as only few vertices should be involved in the regression. This is typically done via a suitable group-sparsity regularization strategy. We then consider a second regularization term $\Omega_2(\beta) = \alpha \sum_{j} \|\beta_j\|^2$, where $\beta_j$ are the coefficients of the vertex $j$. 
Therefore, our predictive model is based on solving the trace regression problem~\eqref{eq:traceRegressionOptim} with the regularization term
\begin{equation}
    \label{eq:regularization}
    \Omega(\beta) = \eta \mathrm{tr}(\beta^{\top} L \beta) / 2 + \alpha \sum_{j} \|\beta_j\|^2~.
\end{equation}
Since $\Omega(\cdot)$  is a convex but non-differentiable function, we use the FISTA algorithm, an accelerated projected gradient descent method, to solve problem~\eqref{eq:traceRegressionOptim}. We use the monotonous version for stability purpose~\cite{beck2009gradient}. 

\renewcommand{\algorithmicrequire}{\textbf{Input:}}
\renewcommand{\algorithmicensure}{\textbf{Output:}}
\begin{algorithm}[!ht]
    \caption{The main pipeline}\label{sect-SSG-algo}
    \begin{algorithmic}
    \Require $\{x_t^i\}_{i=1}^{n} \in \mathbb{R}^{n \times D_{task}}$, $\{x_r^i\}_{i=1}^n\in \mathbb{R}^{n \times D_{rest}}$: a database of two-view observations, $Y \in \mathbb{R}^n$: the regression target.
    \State \textbf{1.} Learn a \ak{multi-view latent representation $z$ from $x_r$ and $x_t$.}
    \State \textbf{2.} From the latent representation built the resulting observation, $\{z_i\}_{i=1}^{n} \in \mathbb{R}^{n_z \times D}$.
    \State \textbf{3.} Apply the trace regression model using $\{Z_i\}_{i=1}^{n} \in \mathbb{R}^{n_z \times D}$ and $y$ as inputs.
    \Ensure {$\beta\in \mathbb{R}^{n_z \times D}$ map used for prediction.}
    \end{algorithmic}
    \label{algo_pipeline}
\end{algorithm}
\section{Experiments}

\label{results}

In this section, we describe the experiments that we performed to assess the relevance and \ak{effectiveness of our method}. We first describe the fMRI dataset that was used, before presenting a thorough quantitative evaluation of both the multi-view representation learning phase and the predictive model itself. Finally, we discuss the neuroscientific relevance of our results. The pipeline used in this section is summarized by Algorithm~\ref{algo_pipeline}.
\subsection{Data set and pre-processing}
\label{preprocessing}
We performed our experiments on the InterTVA dataset \cite{data}, which aims at studying inter-individual differences in voice perception and voice recognition using multi-modal MRI data. We targeted this data set because it offers both task- and rest-fMRI data, as well as a precise behavioral characterization of the 40 subjects available.
More precisely, we exploited: i) the task-fMRI session during which the participants listened passively to 144 vocal and non-vocal sounds, ii) the 12mn long rest-fMRI session, iii) the performance of the subjects in a voice recognition task, measured  \ak{as percentage  the Glasgow Voice
Memory Test (GVMT) \cite{aglieri2017glasgow}}, as well as iv) the high-resolution T1 anatomical image for pre-processing purposes.

Both the task- and rest-fMRI data sets were first corrected for slice-timing and subject's motion (using SPM12, \url{www.fil.ion.ucl.ac.uk/spm}), and the physiological noise components were removed using the Physio Tapas toolbox (\url{www.tnu.ethz.ch/en/software/tapas.html}). The corrected fMRI time-series were projected on a triangulated mesh representing the surface of the cortex (comprising  $20,484$ vertices), that had been extracted from the T1 MRI image using the freesurfer software suite (\url{surfer.nmr.mgh.harvard.edu}). Then, two feature vectors were estimated at each cortical location, i.e. each vertex of the mesh, one for each fMRI protocol. For task-fMRI, the feature vector $x_{t} \in \mathbb{R}^{ D_{task}}$ is the set of amplitudes of the fMRI responses induced by each of the 144 audio stimuli presented to the participant, estimated using a general linear model that included one regressor per stimulus \cite{aglieri_singletrial_2020} (hence, $D_{task}=144$). For rest-fMRI, the feature vector $x_{r} \in \mathbb{R}^{D_{rest}}$ contains the correlation coefficients between the time series of the vertex and each of the 150 average time-series computed in the regions of the Destrieux atlas, available in freesurfer (hence, $D_{rest} = 150$). These two feature vectors are then used as inputs of the multi-view representation learning algorithms described in Section~\ref{method}.A. With a training set size of 36 subjects out of the 40 available (see hereafter), the multimodal autoencoder was trained with $36 \times 20,484$ samples of dimension $D_{concat} = D_{rest} + D_{task} = 294$. The task addressed in our experiments was to predict the GVMT score of each participant from the full-brain information carried in these connectivity and activation feature sets, using the trace regression model trained on 36 subjects.

  \subsection{Deep autoencoders implementations}
 We used different types of autoencoders: linear and non linear monomodal (simple) autoencoder (AE),   multimodal deep autoencoder (MDAE). For all the models, we investigated several encoding dimensions from 2 to 100 ($enc \in [2,..,100]$). Table \ref{architecture} reports the different architectures of AE and MDAE models.
 
\begin{table}[!ht]
    \caption{Details on architectures of investigated AE/MDAE models. The models are built by stacking fully connected layers whose dimensions are given from input-layer to output layer, with \emph{enc} being the dimension of the encoding. For instance [$D_{concat}$, $enc$, $D_{concat}$] stands for a one hidden layer architecture taking as input (and learned to output) the concatenation of $x_r$ and $x_t$, while projecting through a hidden layer of size \emph{enc}.}
     \label{architecture}
    \centering
\begin{tabular}{L{2.9cm}|L{5.35cm}}
Model (Number of hidden layers)& Architecture\\
\hline\hline
   & [$D_{task}$, $enc$, $D_{task}$] \\
  AE (one layer)  &   [$D_{rest}$, $enc$, $D_{rest}$]\\
   &  [$D_{concat}$, $enc$, $D_{concat}$] \\
\hline
 & [$D_{task}$, 120, $enc$, 120,  $D_{task}$] \\
    &   [$D_{task}$, 130, $enc$, 130,  $D_{task}$]\\
  MDAE/AE (two layers) &  [$D_{rest}$, 120, $enc$, 120,  $D_{rest}$]\\
   &  [$D_{rest}$, 130, $enc$, 130,  $D_{rest}$] \\
   &  [$D_{concat}$, 150, $enc$, 150,  $D_{concat}$] \\
   & [$D_{concat}$, 200, $enc$, 200,  $D_{concat}$] \\
\hline
& [$D_{task}$, 140, 120, $enc$, 120, 140,  $D_{task}$]\\
& [$D_{task}$, 140, 130, $enc$, 130,  140, $D_{task}$]\\
MDAE/AE (three layers) & [$D_{rest}$, 140, 120, $enc$, 120, 140,  $D_{rest}$]\\
& [$D_{rest}$, 140, 130, $enc$, 130,  140, $D_{rest}$]\\
& [$D_{concat}$, 250, 150, $enc$, 150, 250,  $D_{concat}$]\\ & [$D_{concat}$, 200, 130, $enc$, 130,  200, $D_{concat}$]\\
\hline\hline
\end{tabular}
\end{table}
For all the network models, we tested different pairs of activation functions for the hidden layers and output layer, respectively: (\textit{linear, linear}), (\textit{linear, sigmoid}), (\textit{relu, linear}), and (\textit{relu, sigmoid}).

All different autoencoders models were implemented using the \textit{keras} toolkit. Training was performed using the Adam optimizer \cite{kingma2014adam}, with a learning rate ($lr$) equals to $10^{-3}$ over $300$ epochs and a batch size of $500$ samples.
\subsection{\ak{Benchmarked} representation learning methods}
We performed a comparative study between several representation learning models, including the two MDAE models (concatenated inputs, and concatenated latent representations), standard AE, principal component analysis (PCA) \cite{bishop2006pattern}, independent component analysis (ICA) \cite{hyvarinen2000independent}, deep canonically correlated autoencoders (DCCAE) \cite{wang2015deep}. 
\subsubsection{PCA} \ak{as baseline, we investigated the use of PCA to reduce the dimensionality of multimodal fMRI data. PCA seeks an optimal linear orthogonal transformation that provides a new coordinate system, i.e., the latent space, in which basis vectors follow modes of greatest variance in the original fMRI data \cite{bishop2006pattern}. }
\subsubsection{ICA} \ak{an extension of PCA technique, which aims to optimize higher-order statistics such as kurtosis. Usually, it is used as a computational method for separating a multivariate signals into additive sub-components \cite{hyvarinen2000independent}.}
\subsubsection{DCCAE} the aim is to optimize the combination of canonical correlation between the learned bottleneck representations (latent space) and the reconstruction errors of the autoencoders \cite{wang2015deep}. In fact, it consists of two autoencoders on top of two deep neural networks.
  
 \subsection{\ak{Evaluation scheme}}

In order to train the different models, we  perform a 10-fold cross-validation, i.e., for each fold, we used $36$ subjects for training and $4$ subjects for testing. Moreover, we compared all representation learning methods (AE, MDAE, DCCAE, PCA, and ICA) based on the prediction error by computing the MSE between the true behavioral score noted by $y$  and the predicted behavioral score estimated by the trace regression model noted by $\hat{y}$. Therefore, we tested all methods on 10-fold cross-validation, and we reported the average MSE for each method. Moreover, we computed the average R-squared (coefficient of determination) noted by $R^{2}$, which is the fraction by which the variance of the errors is less than the variance of the dependent variable.  It can be defined as follows:
\begin{equation}
    R^{2}=  1- \frac{\sum_{i=1}^{N}(y_{i} - \hat{y}_{i})^2}{\sum_{i=1}^{N}(y_{i} - \bar{y}_{i})^2}
\end{equation}
\ak{where $\hat{y}_{i}$ is the predicted behavioral score and the $\bar{y}_{i}$ is the mean of behavioral scores}. The $R^{2}$ indicates how well the model predictions approximate the true values. 
\begin{figure}[!ht]
    \centering
    \includegraphics[scale=0.35]{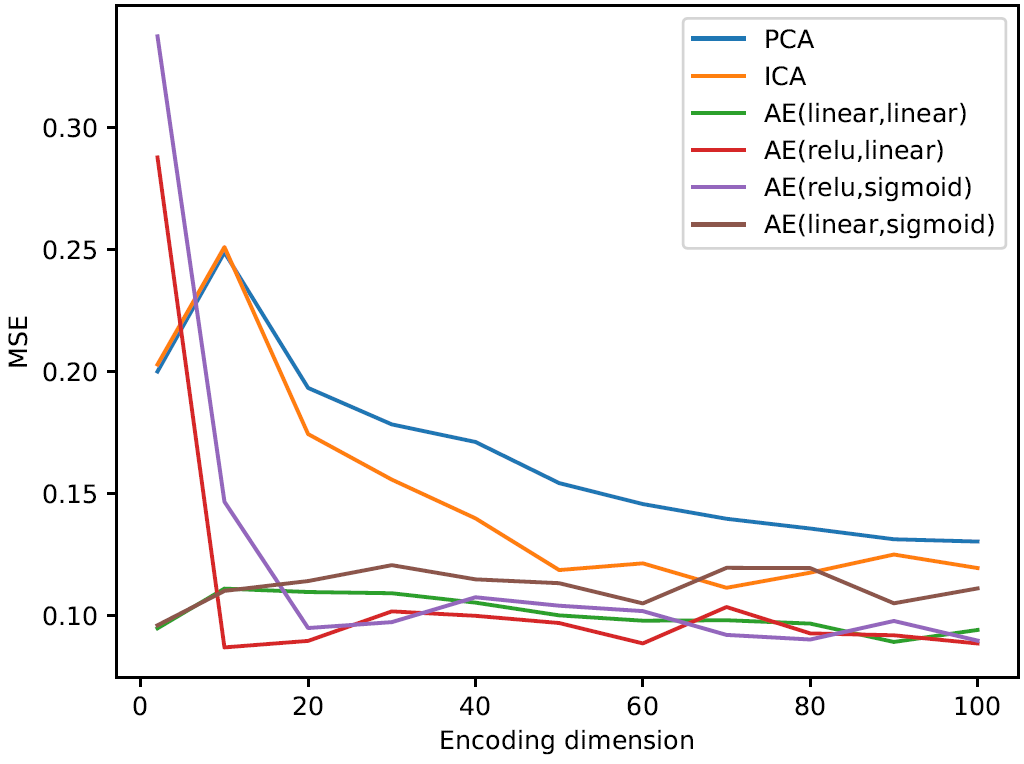}
    \caption{Average MSE  versus encoding dimension across 10-fold cross validation using task-based fMRI data }
    \label{monomodal_regression}
\end{figure}
\begin{table*}[!ht]
    \caption{\ak{Best average MSE} and $R^{2}$  ($\pm$ standard error)  using monomodal fMRI data  estimated on trace regression model after regularization with spatial constraint ($L1)$ and parsimony ($L2$) based on PCA, ICA, and different AE models (3 layers is the best architecture: [$D_{task}$, 140, 120, $enc$, 120, 140,  $D_{task}$] and [$D_{rest}$, 140, 120, $enc$, 120, 140,  $D_{rest}$]). }
     \label{regression_mono}
    \centering
\begin{tabular}{l|c|c||c|c} 
  \hline
     & \multicolumn{2}{c||}{task-fMRI} & \multicolumn{2}{c}{rest-fMRI} \\
    \hline
    \cline{2-5} 
      & Average MSE & Average $R^{2}$ & Average MSE & Average $R^{2}$  \\
    \hline\hline
    PCA& 0.152  ($\pm$ 0.0035) & 0.129 ($\pm$ 0.0051) & 0.345  ($\pm$ 0.0082) & 0.071 ($\pm$ 0.0047)\\
    \hline
    ICA& 0.137 ($\pm$ 0.0094) & 0.162 ($\pm$ 0.0085) &0.345  ($\pm$ 0.0032) & 0.073  ($\pm$ 0.0124)\\
    \hline
    AE (\textit{linear, linear}) & 0.091 ($\pm$ 0.0071) & 0.239 ($\pm$ 0.0023)& \textbf{0.340}  ($\pm$ 0.0195)& \textbf{0.094} ($\pm$ 0.0025) \\
    \hline
    AE (\textit{linear, sigmoid}) & 0.093 ($\pm$ 0.0026) & 0.201 ($\pm$ 0.0096) &0.344 ($\pm$ 0.0088) &  0.081 ($\pm$ 0.0040) \\
    \hline
    AE (\textit{relu, linear}) &  \textbf{0.082} ($\pm$ 0.0088)& \textbf{0.259} ($\pm$ 0.0075)& 0.344  ($\pm$ 0.0163)& 0.086 ($\pm$ 0.0019)  \\
    \hline
    AE (\textit{relu, sigmoid}) & 0.102 ($\pm$ 0.0053) & 0.112 ($\pm$ 0.0029)& 0.342  ($\pm$ 0.0179)& 0.089  ($\pm$ 0.0201)\\
    \hline
    Raw fMRI data &  0.632 ($\pm$ 0.0289) &-2.703 ($\pm$ 0.3055) &0.905  ($\pm$ 0.0179)  & -8.384 ($\pm$ 0.4241)\\
\end{tabular}
\end{table*}
\begin{table*}[!ht]
    \caption{\ak{Best average MSE} and $R^{2}$  ($\pm$ standard error)  using multimodal fMRI data  estimated on trace regression model after regularization with spatial constraint ($L1)$ and parsimony ($L2$) based on PCA, ICA, DCCAE, and different AE/MDAE models (3 layers is the best architecture: [$D_{concat}$, 200, 130, $enc$, 130,  200, $D_{concat}$]). }
     \label{regression_multi_table}
    \centering
\begin{tabular}{l|c|c||c|c} 
  \hline
     & \multicolumn{2}{c}{Concatenated inputs (task+rest-fMRI) } & \multicolumn{2}{c}{Concatenated latent representations ($z_{t} + z_{r}$) }\\
    \hline
    \cline{2-4} 
      & Average MSE & Average $R^{2}$ & Average MSE & Average $R^{2}$\\
    \hline\hline
    PCA& 0.091 ($\pm$ 0.0021)  &0.239 ($\pm$ 0.0048) &N/A&N/A \\
    \hline
    ICA& 0.092 ($\pm$ 0.0109) &  0.223 ($\pm$ 0.0090)&N/A&N/A\\
    \hline
    AE/MDAE (\textit{linear, linear}) &0.083 ($\pm$ 0.0067)  & 0.254 ($\pm$ 0.0104)& 0.065 ($\pm$ 0.0058) & 0.273 ($\pm$ 0.0033) \\
    \hline
    AE/MDAE (\textit{linear, sigmoid}) & 0.083 ($\pm$ 0.0089)  &0.249 ($\pm$ 0.0036) & 0.079 ($\pm$ 0.0124) & 0.249 ($\pm$ 0.0204)  \\
    \hline
    AE/MDAE (\textit{relu, linear}) & \textbf{0.080} ($\pm$ 0.0032)  & \textbf{0.262} ($\pm$ 0.0136) & \textbf{0.062} ($\pm$ 0.0095)  & \textbf{0.282} ($\pm$ 0.0044) \\
    \hline
    AE/MDAE (\textit{relu, sigmoid}) & \textbf{0.080} ($\pm$ 0.0074)  &0.260 ($\pm$ 0.0281)& 0.073 ($\pm$ 0.0086)  &0.254 ($\pm$ 0.0051)\\
     \hline
        DCCAE &  N/A & N/A&$0.126$ $(\pm 0.0075)$ & $0.183$ $(\pm 0.0093)$ \\
    \hline
    Raw fMRI data ($x_{t} + x_{r}$) & N/A & N/A&-&- \\
\end{tabular}
\end{table*}
\begin{figure*}[!ht]
    \centering
    \includegraphics[scale=0.19]{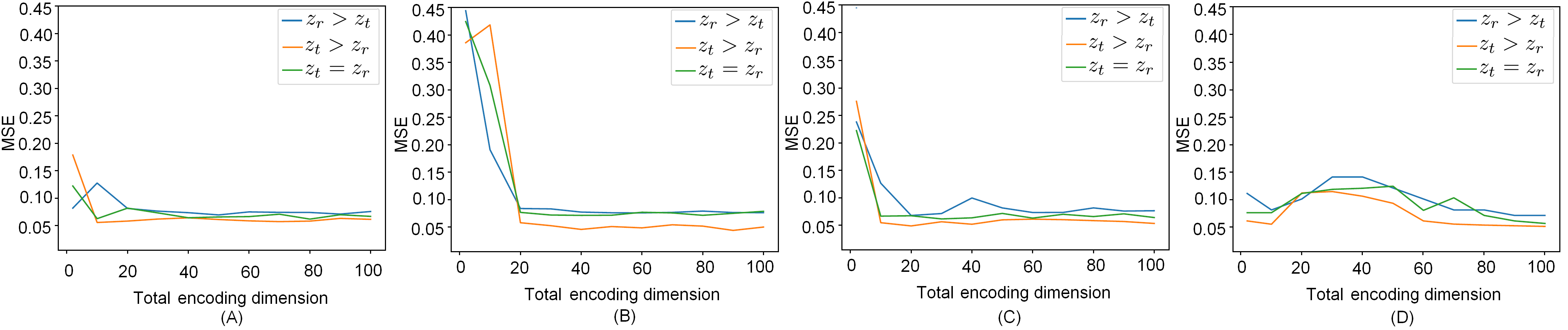}
    \caption{Comparison of the average MSE (over the 10-fold cross-validation) for different relative sizes given to the task- and rest-fMRI modalities in the bottleneck layer of the MDAE (larger size for rest-fMRI: $z_{r} > z_{t}$; larger size for task-fMRI: $z_{t} > z_{r}$; equal sizes: $z_{t} = z_{r}$ (A). MDAE (\textit{linear, linear}),  (B). MDAE (\textit{linear, sigmoid}),  (C). MDAE (\textit{relu, linear}), (D). MDAE (\textit{relu, sigmoid}). Overall, the performances are higher when $z_{t} > z_{r}$.}
    \label{contribution_modality}
\end{figure*}
\begin{figure*}[t!]
    \centering
    \includegraphics[scale=0.51]{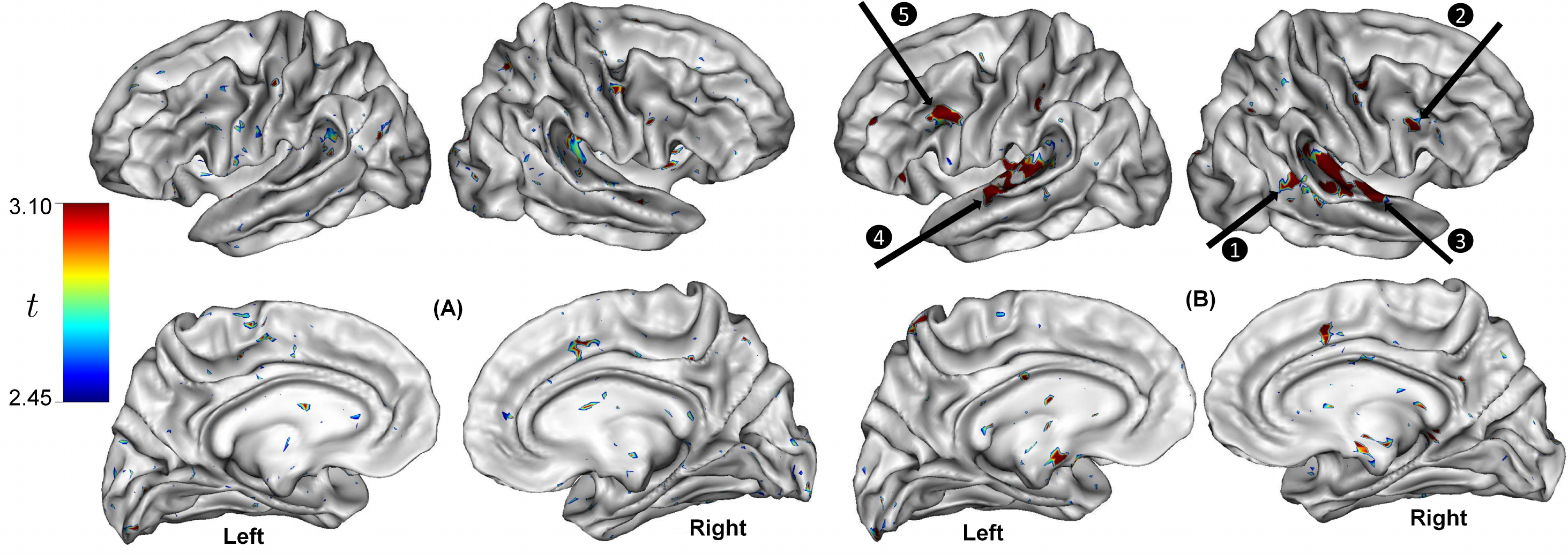}
    \caption{Average weight maps estimated using our task+rest fMRI-based predictive model of individual differences, thresholded after a test for statistical significance ($t>2.45$, $p < 0.005$). Regions of non zero average weight appear in color. (A) With the model using concatenated inputs in the MDAE (\textit{relu, sigmoid}), only small scattered regions are detected. (B) When using the concatenated latent representations in the MDAE (\textit{relu, sigmoid}), several larger regions are detected: regions tagged 3 and 4 are along the superior temporal gyrus bilaterally, region 1 is located in the fundus of the right superior temporal sulcus, regions 2 and 5 are in the inferior frontal gyrus. All these regions perfectly match the neuroscientific literature \cite{belin_voice-selective_2000, pernet_human_2015,aglieri_functional_2018}. }
    \label{Average}
\end{figure*}
\subsection{Results: quantitative evaluation of the predictive model}
In this section, we present the  results obtained by our predictive model. The trace regression model~\eqref{eq:traceRegressionModel} is estimated on the latent representation $Z \in \mathbb{R}^{n \times d}$. For the \ak{spatial constraints}, we use the Laplacian matrix of the graph given by the triangulated cortical mesh. \ak{Indeed, we use the null vector as initialization in order to catch only useful features. This spatial regularization makes it possible to favor the nodes having the same value as their neighbors. }\ak{We empirically found by cross-validation $\alpha$ and $\eta$ respectively equals to $5e^{-4}$ and $1e^{-3}$ are the best hyper-parameters values}. Moreover, in order to benchmark our approach, we also applied \ak{the trace regression model} on the raw fMRI data. Hence, we perform two main experimentation for the evaluation of the predictive model, using unimodal and multimodal fMRI data.
\subsubsection{Prediction performances using monomodal fMRI data}
In order to study individual differences from such a dataset, a neuroscientist would classically use the single data modality that relates the most closely to nature of the task used to measure the behavioral performance, hence task-fMRI (listening to voices in task-fMRI, recognizing voices in the GVMT behavioral score).
We therefore provide a comparative study between representation learning methods, including PCA, ICA, linear AE, and non-linear AE with different pairs activation functions using monomodal task-fMRI data. The aim, is to study the influence of the nature of the representation learning models on the regression task. Figure \ref{monomodal_regression} reports the \ak{average MSE versus encoding dimension using task-fMRI data}. We can notice that the best MSE for task-fMRI is obtained using the AE\textit{(relu, linear) } model, which is equal to $0.09$, with an encoding dimension of 10. We can also observe that all AE models are effective with few dimensions ($enc$ is between 10 and 20 features), while the other models require using many more dimensions to reach their best performance, which demonstrates the efficiency of the AE models at compressing the data efficiently in this context. For the sake of exhaustivity, we also compare the performances of the models estimated from monomodal task-fMRI data and from monomodal rest-fMRI data:  Table~\ref{regression_mono} includes the best (across encoding dimension) average MSE and $R^{2}$ over the 10-fold cross validation, for all representation learning models.  
The performances obtained from monomodal rest-fMRI data are far worst than the ones obtained for monomodal task-fMRI data. This was to be expected since the task-fMRI data involved a cognitive task that was very close to the one performed during the behavioral GVMT test, while the rest-fMRI data only contains non-specific information of the whole-brain functional connectivity.
Moreover, the performances were also a lot better when using a representation learning method than when using the raw data (last row of Table~\ref{regression_mono}), both in terms of MSE and $R^{2}$. This demonstrates the added value of using such representation learning approach for this task.
\subsubsection{Multimodal representation learning: combining task- and rest-fMRI} In this section, we compare the different implementations of the two multimodal autoencoder introduced in Section~\ref{method}.A, i.e. when the task- and rest-fMRI inputs are combined to learn a multi-view representation. We compare them, again, using MSE and $R^{2}$. Hence, we report in Table~\ref{regression_multi_table} the best (across encoding dimensions) average MSE and $R^{2}$ over 10-fold cross validation using combined fMRI data for each method.  
The best performances obtained using the first architecture (AE, concatenation of the inputs, left part of Table~\ref{regression_multi_table}) are MSE = $0.080$ and $R^{2}=0.262$ for the AE(\textit{relu, linear}) model, which are barely better than the performances obtained from the monomodal model based on task-fMRI.
However, we obtain an important gain of performances with our second multi-view architecture (MDAE, concatenation of the latent representations): all MDAE models actually outperform the monomodal models, the multimodal models based on concatenated inputs, as well as the deep canonically correlated autoencoder multimodal model (DCCAE). The best model is the MDAE(\textit{relu, linear}) model, with a MSE = $0.062$ and $R^{2}=0.282$.
This clearly demonstrates the added value of fusing task- and rest-fMRI in order to study inter-individual differences, and that a multimodal autoencoder that operates a concatenation of the latent representation is an efficient representation learning scheme for this objective.

\subsubsection{Influence of the relative size of the rest- and task-fMRI in the bottleneck layer of the MDAE}

Having demonstrated the effectiveness of our approach, the objective of this section is to attempt to differentially weight the two modalities in order to further improve the performances obtained using our MDAE with concatenated latent representations. 
To do so, we vary the sizes of the latent representations $z_t$ and $z_r$ of each modality for a given total encoding dimension. We present three configurations where each configuration corresponds to different sizes of the bottleneck layers $z_{t}$ and $z_{r}$. Thereby, in the first configuration,  the size of $z_{r} > z_{t}$. In the second configuration, the size of $z_{r} < z_{t}$, and in the last configuration the sizes of $z_{r}$ and $z_{t}$ are equal ($z_{r} = z_{t}$).
Figure~\ref{contribution_modality} reports the average MSE computed over the 10-fold cross-validation using the four MDAE models with different pairs of activation functions, which are the four best models identified previously (see Table~\ref{regression_multi_table}). Using these three configurations, we aim here at evaluating the importance of each modality for the regression task.
Overall, the best MSEs are obtained for the second configuration, i.e. when $z_{t} > z_{r}$, showing that the task-fMRI provides more information than the rest-fMRI data. For instance,  the best MSE is equal to $0.050$ with MDAE(\textit{relu, sigmoid}) and the $R^{2}$ is of $0.294$ $(\pm 0.0048)$,  where 8 features extracted from task-fMRI and 2 features extracted from rest-fMRI. Furthermore, for the MDAE(\textit{linear, linear}) model, the best average MSE and $R^{2}$ are obtained with an encoding dimension of  10, where again $z_{t} > z_{r}$. For the MDAE(\textit{relu, linear}) model, 20 features allow to get a higher prediction with 15 features from task-fMRI and 5 features from rest-fMRI. For the MDAE (\textit{linear, sigmoid}) model, the best MSE is equal to $0.052$ where 12 features extracted from task-fMRI and 8 features from rest-fMRI (20 features). We can conclude then that the task-fMRI contributes more than the rest-fMRI, where the MSE is very low and the prediction accuracy is sufficient. Moreover, we have demonstrated that the concatenation of bottleneck layer is more suitable than the concatenation of inputs.

\subsection{Neuroscientific relevance of the model}

 We now examine the neuroscientific relevance of our results.
Lacking a ground truth, we compare our results qualitatively with state of the art knowledge extracted from
the neuroscientific literature \cite{belin_voice-selective_2000, pernet_human_2015,aglieri_functional_2018}:
knowing the task performed by the subject (i.e the passive listening of vocal sounds), we can expect to see a very focal network of
brain regions located bilaterally in the temporal lobe (along the superior temporal gyrus and sulcus \cite{belin_voice-selective_2000, pernet_human_2015}),
as well as regions in the frontal lobe (in the pre-central and inferior frontal gyri \cite{aglieri_functional_2018}).
For this, we present on Fig.~\ref{Average} the average $\hat{\beta}$ weight maps estimated using the trace regression,
for the multi-modal representations obtained with either concatenated inputs or concatenated latent representations.
To ease the interpretation, these maps are presented overlayed on the three-dimensional cortical mesh, after performing a statistical test to extract significant clusters of non-zero average weights (t-test, performed in SPM12; thresholded at $p<0.005$, i.e $t>2.45$).
The weight maps estimated using the concatenation of the inputs (on the left) present very few regions of small sizes where the average weight is significantly non zero, scattered all over the cortex.
In contrast, the multi-view autoencoder based on concatenated latent representations (on the right) yields several larger significant regions which closely correspond to the expectations described previously (see arrows on Fig.~\ref{Average}).
Remarkably, it allows detecting significant regions in the frontal lobe bilaterally, regions that are known to be hard to detect (e.g. 92 subjects were used in~\cite{aglieri_functional_2018} whereas only 39 were available in the present study).
This might reflect a gain in statistical power that could be induced by an improved robustness of the information present in the latent representation of the fused task- and rest-fMRI data. Further experimental validation on other datasets should be conducted to confirm this potential gain in statistical power.

\section{Conclusion}
\label{conclusion}

In this paper, we introduced a novel machine learning method which aims at mapping individual differences in cortical architecture using multi-view representation learning. Our method seeks to fuse task- and rest-fMRI in order to exploit the complementary information they offer about brain activation and connectivity respectively.
To do so, a deep multi-view autoencoder was designed to fuse the two fMRI modalities, yielding a compressed joint representation space within which
a trace regression model is developed.
This model allows predicting the behavioral performances of new patients. Our experimental results demonstrate the ability of the proposed method to outperform competitive approaches and produce interpretable and biologically plausible results with a potential gain in statistical power. In the future, the proposed method can be extended by introducing a graph convolutional network model incorporating  the 3-D cortical mesh in order to preserve the spatial structure, but also by adding extra data modalities that can characterize structural connectivity or the local folding pattern of the cortex.

\section*{Acknowledgment}
This work was carried out within the \textit{Institut Convergence ILCB} (ANR-16-CONV-0002) and with access to the high performance computing resources of the \textit{Centre de Calcul Intensif d’Aix-Marseille}. It was also funded in part by the French \textit{Agence Nationale de la Recherche} (grants ANR-15-CE23-0026 and ANR-16-CE23-0006).

\bibliographystyle{IEEEtran}
\bibliography{biblio}

\end{document}